\documentclass[11pt]{article}

\usepackage{moreverb}

\usepackage[colorlinks,bookmarksopen,bookmarksnumbered,citecolor=red,urlcolor=red]{hyperref}
\usepackage[numbers]{natbib}
\usepackage{bbm}
\usepackage{epstopdf}
\usepackage{subcaption}
\usepackage{amsmath}
\usepackage{graphicx}
\usepackage{fullpage}

\def \s2{\sigma^2}
\def \hs2{\hat{\sigma}^2}
\def \siga2{\sigma_{\alpha}^2}
\def \sige2{\sigma_{\epsilon}^2}
\def \sig2{\sigma^2}

\newcommand{\bx}{\mbox{\boldmath $x$}}

\newcommand{\bX}{\mbox{\boldmath $X$}}

\newcommand{\bY}{\mbox{\boldmath $Y$}}
\newcommand{\bmy}{\mbox{\boldmath $y$}}

\newcommand{\beq}{\begin{equation}}       
\newcommand{\eeq}{\end{equation}}       
\newcommand{\beqn}{\begin{eqnarray}}
\newcommand{\eeqn}{\end{eqnarray}}

\newcommand{\bbeta}{\mbox{\boldmath $\beta$}}

\newcommand{\convto}{\rightarrow}

\newcommand{\indicatorBig}[1]{\mathbbm{1}{\left[ {#1} \right] }}

\newcommand{\scriptT}{\mathcal{T}}

\newcommand{\given}{\ | \ }

\newcommand\BibTeX{{\rmfamily B\kern-.05em \textsc{i\kern-.025em b}\kern-.08em
T\kern-.1667em\lower.7ex\hbox{E}\kern-.125emX}}

\begin{document}

\title{A Naive Bayes machine learning approach to risk prediction using censored, time-to-event data}
\author{J. Wolfson\footnotemark[1], S. Bandyopadhyay\footnotemark[2], M. Elidrisi\footnotemark[2], G. Vazquez-Benitez\footnotemark[4],\\ D. Musgrove\footnotemark[1], G. Adomavicius\footnotemark[3], P. Johnson\footnotemark[3], and P. O'Connor\footnotemark[4]}

\maketitle

\begin{abstract}
Predicting an individual's risk of experiencing a future clinical outcome is a statistical task with important consequences for both practicing clinicians and public health experts.  Modern observational databases such as electronic health records (EHRs) provide an alternative to the longitudinal cohort studies traditionally used to construct risk models, bringing with them both opportunities and challenges. Large sample sizes and detailed covariate histories enable the use of sophisticated machine learning techniques to uncover complex associations and interactions, but observational databases are often ``messy,'' with high levels of missing data and incomplete patient follow-up. In this paper, we propose an adaptation of the well-known Naive Bayes (NB) machine learning approach for classification to time-to-event outcomes subject to censoring. We compare the predictive performance of our method to the Cox proportional hazards model which is commonly used for risk prediction in healthcare populations, and illustrate its application to prediction of cardiovascular risk using an EHR dataset from a large Midwest integrated healthcare system.\\
\end{abstract}


\footnotetext[1]{Division of Biostatistics, University of Minnesota, Minneapolis, MN. Correspondence to: julianw@umn.edu}
\footnotetext[2]{Department of Computer Science and Engineering, University of Minnesota, Minneapolis, MN}
\footnotetext[3]{Department of Information and Decision Sciences, Carlson School of Management, University of Minnesota, Minneapolis, MN}
\footnotetext[4]{HealthPartners Institute for Education and Research, Minneapolis, MN}

\section{Introduction}

The task of predicting the occurrence of future clinical events is central to many areas of medical practice. ``Risk calculators'', which make personalized risk predictions based on individual characteristics, are now commonly employed for outcomes such as heart attack, stroke, and diabetes. A common feature of many of these risk calculators is that their predictions are based on data from cohort studies. For example, the Framingham Risk Score (FRS) \cite{FRS,DAgostino2001} for predicting the occurrence of cardiovascular disease (CVD) is based on data from the Framingham Heart Study \cite{FHS}, a cohort study which has been ongoing since 1948. Many calculators, including the FRS, focus on predicting five- and ten-year risk, i.e., the probability of having a first CVD-defining event (coronary heart disease, stroke, or other CVD-related death) in the next five or ten years. Unfortunately, the cost of subject acquisition and long-term follow-up restricts the size of these studies; for example, total enrollment in the Framingham Heart Study across three generational cohorts is slightly less than 15,000 \cite{FHSwebsite}, and a recent study of a subset of 8,491 FHS participants \cite{DAgostino2008} reported 1,174 incident CVD events over 12 years of follow-up.  As a result, longitudinal cohort studies may be limited in their ability to describe unique features of contemporary patient populations, and in their capacity to predict risk over shorter time horizons (eg. 1 year, 5 years) when the event rate is low. 

An alternative data source for risk prediction is electronic health record (EHR) data, which is increasingly available on entire populations under care within health insurance systems. A typical EHR data capture covers a fixed time period, often several years, during which subjects may enroll in and/or disenroll from the health insurance plan.  EHR data often contain hundreds of thousands of records on enrolled subjects from multiple sources including electronic medical records, claims and prescription data, and state and national databases.  Since the data are recorded longitudinally, they are well-suited to obtaining risk estimates, but their complexity, dimensionality, and (often) high levels of missing data challenge standard statistical methods.

Risk engines for a variety of clinical outcomes have traditionally been developed by fitting survival regression models such as the Cox proportional hazards model \cite{Cox1972} or the accelerated failure time model \cite{Buckley1979} to censored time-to-event data from the aforementioned longitudinal studies. Because regression-based models yield estimates of the effects of each covariate in the model, they are useful not only as a predictive tool, but also for understanding the relative influence of covariates on risk. However, the drawback of fitting simple regression models is that they often lack the flexibility to capture non-linear covariate effects and interactions. If predictive performance and not model interpretability is the central concern, then more flexible machine learning techniques may yield improved risk engines by capturing complex relationships and between covariates and risk \cite{Chia2012}.

In this paper, we present Censored Naive Bayes (CNB), a machine learning approach to predicting risk using time-to-event data subject to censoring.  CNB is an extension of Naive Bayes (NB), a popular technique typically applied to classification problems with categorical outcomes. CNB is nonparametric with respect to the underlying distribution of event times, and models the marginal covariate distributions in a flexible manner. We begin by describing our setting of interest, and describe an extension of the usual Naive Bayes technique allowing estimation of the conditional survivor function. The estimated survivor function can be used to predict risk over any specified time interval. We compare our proposed CNB technique to a Cox proportional hazards (CPH) based approach in a variety of simulated scenarios, and illustrate the application of our method to cardiovascular risk prediction using longitudinal electronic health record data from a large Midwest integrated healthcare system.


%
\section{Naive Bayes for binary outcomes}

Naive Bayes is a machine learning technique that has been widely applied to classification problems \cite{Domingos1996,Domingos1997,Hand2001,Singh2005,Maimon2005}. In this section, we describe the basic NB technique for estimating the probability of a binary outcome. In the next section, we discuss our extension of the NB to time-to-event data with censored outcomes.

Consider the task of estimating the probability of occurrence of an event $E$ over a fixed time period $\tau$, based on individual characteristics $\bX = (X_1, \dots, X_p)$ which are measured at some well-defined baseline time. For the moment, we will assume that we have data on $n$ subjects who have had $\bX$ measured at baseline and been followed for $\tau$ time units to observe whether or not $E$ occurred ($E=1 \Rightarrow$ event occurrence). The target of of estimation is $P(E=1 \given \bX)$.


While regression-based approaches derive a likelihood or estimating function by directly modeling $E$ as a function of $\bX$, an alternative is to proceed indirectly by using Bayes Rule, writing
\[
P(E = 1 \given \bX) = \frac{P(\bX \given E=1) P(E=1)}{ P(\bX) } = \frac{P(\bX \given E=1) P(E=1)}{ \sum_{e=0,1} P(\bX \given E=e)P(E=e) }.
\]
In general, it may be difficult to model the joint distribution of $\bX | E$, particularly if $p$ is large. Researchers have therefore proposed to simplify estimation by assuming parsimonious dependence structures for $\bX \given E$. The \textit{Bayesian network} \cite{Pearl1986} approach consists of partitioning $\bX$ into low-dimensional, mutually independent subsets on the basis of a directed acyclic graph (DAG) encoding conditional independence relationships between covariates. The most extreme version of this approach, termed \textit{Naive Bayes}, makes the dramatic simplifying assumption that the covariates $X_1, X_2, \dots, X_p$ constituting $\bX$ are independent, given $E$, so that
\begin{equation}
P(E = 1 \given \bX) = \frac{ \prod_{j=1}^p P(X_j \given E=1) P(E=1)  }{ \sum_{e=0,1} \prod_{j=1}^p P(X_j \given E=e)P(E=e) }.
\label{eq:NBbin}
\end{equation}
A number of parametric and semi-parametric approaches to modeling the univariate covariate distributions are possible and have been described elsewhere \cite{Domingos1997,John1995}; one common assumption in classification problems is that the individual covariate distributions can be represented using mixtures of Normal distributions. Estimation proceeds by computing the maximum likelihood parameter estimates of each covariate distribution (conditional on $E=0,1$), and plugging the resulting estimates into \eqref{eq:NBbin}. $P(E=e)$ is typically estimated as the sample proportion of subjects with $E=e$. 

Even though the Naive Bayes assumption is often incorrect, NB has been found to perform quite well in a variety of classification tasks, including when covariates are highly correlated \cite{Domingos1996,Domingos1997,Titterington1981,Mani1997,Russek1983}. NB is also known to be poorly calibrated in some settings, i.e., conditional probability estimates are biased though they are highly correlated with the true probabilities so that classification performance remains acceptable \cite{Zadrozny2001}. NB easily accommodates problems with large numbers of covariates, can be computed quickly for large samples, and implicitly handles datasets with missing covariate values. The problem of NB calibration is addressed at greater length in the Simulations and Discussion sections. 

\section{Censored Naive Bayes}

\subsection{Limtations of the binary-outcomes approach}

In many studies, some participants will have incomplete data because of loss-to-follow-up or censoring. The EHR dataset motivating this work consists of patient records captured between 2002 and 2009; over this time period, some patients who contributed data in 2002 disenrolled prior to 2009, and other patients enrolled and began contributing data after 2002. If $E$ denotes the indicator that a subject experiences a cardiovascular event within 5 years of his/her ``index visit'' (a term we define more formally in the next section), then subjects who were followed for less than five years and who did not experience an event have $E$ undefined. One approach is to discard such subjects (see, e.g., \cite{Sierra1998,Blanco2005,Larranaga1997}), but when a substantial fraction of subjects do not have five years of follow-up this strategy may lead to biased probability estimates since the underlying failure times are right-truncated. The typical classification-based approach suffers from other drawbacks as well: For example, changing the time horizon over which one wants to make predictions (e.g., from 5 years to 3 years) requires recalculating the outcomes and refitting the model, which may be impractical if the NB is to be used as the basis of a risk calculator in clinical practice. \cite{Zupan2000} and \cite{Stajduhar2010} have proposed approaches which impute the event times of censored observations in a reasonable but somewhat ad hoc manner. \cite{Stajduhar2012} adopts a more principled likelihood-based approach to imputing event times, but their imputation technique may perform poorly if the assumed parametric distribution of event times is incorrect.

\subsection{Notation and setup}

Consider a longitudinal study of $K$ years duration, from which data on $n$ independent subjects are available. Subjects may enter the study at any time between the beginning of the study and year $K$, and follow-up can end when an event occurs, when the study ends in year $K$, or when a subject drops out of the study prior to year $K$. For each subject $i$ define $T_i$ as the time between some index time $t_{i,0}$ and the occurrence of an event (event time), and $C_i$ as the time between $t_{i,0}$ and the end of follow-up due to study termination or drop-out (censoring time). In our data application example, the index time $t_{i,0}$ corresponds to the first clinic visit following the initial 12 months of continuous enrollment in the healthcare system. For all subjects $i=1, \dots, n$, we observe data $(O_i, \delta_i, \bX_i)$, where $O_i$ is the observed time $\min(T_i, C_i)$ , $\delta_i = \indicatorBig{T_i \leq C_i}$, and $\bX_i = \{ X_{i1}, \dots, X_{ip} \}$ is a $p$-vector of continuous-valued covariates available at $t_{i,0}$. The target of inference is now $S_{\bX}(t) = P(T \geq t \given \bX)$, the survivor function conditional on $\bX$. In what follows, we will use $f(A|B)$ as shorthand for the conditional pmf or density of $A$, given $B$.

By Bayes Theorem, we can write
\begin{equation}
S_{\bX}(t) \equiv P(T \geq t \given \bX) = \frac{P(T \geq t) P(\bX \given T \geq t)}{P(\bX)} = \frac{P(T \geq t) P(\bX \given T \geq t)}{P(\bX \given T \geq t)P(T \geq t) + P(\bX \given T < t)P(T < t)}
\label{eq:bayesthm}
\end{equation}
In this context, our ``naiveness'' assumption is that covariates are independent conditional on the events $\{T \geq t\}$ and $\{T < t\}$, for all $t$. We therefore rewrite \eqref{eq:bayesthm} as
\begin{equation}
S_{\bX}(t) = P(T \geq t) \left[ \frac{ \prod_{j=1}^p f(X_j \given T \geq t) }{ \prod_{j=1}^p f(X_j \given T \geq t) P(T \geq t) + \prod_{j=1}^p f(X_j \given T < t) P(T < t) } \right].
\label{eq:naivebayes}
\end{equation}
The key components which require estimation are therefore the marginal survivor function $S(t) \equiv P(T \geq t)$ and the covariate densities $f_j^\geq(x;t) \equiv dP(X_j \leq x \given T \geq t)/dx$ and $f_j^<(x;t) \equiv dP(X_j \leq x \given T < t)/dx$. The following sections describe an approach to estimating these components and hence the resulting conditional survivor function $S_{\bX}(t)$.

\subsection{Estimation of the marginal survivor function $S(t)$}

We propose to estimate $S(t)$ non-parametrically via the Kaplan-Meier estimator \cite{Kaplan1958}. Let the observed event times be $t_1 < t_2 < \dots < t_K$.  For $t_k < t \leq t_{k+1}$, we estimate
\[
\hat{S}_{KM}(t) = \prod_{m=1}^k \left( 1 - \frac{ e_m }{ r_m } \right)
\]
where $e_m$ and $r_m$ respectively give the number of events and individuals at risk at time $t_m$, as follows:
\begin{align*}
e_m &= \sum_{i=1}^n \delta_i \indicatorBig{O_i = t_m}, \\
r_m &= \sum_{i=1}^n \indicatorBig{O_i \geq t_m}.
\end{align*}

Since the Kaplan-Meier estimator makes no assumptions about the form of $S(t)$, it is ideally suited to a machine learning situation where little information may be available to guide parametric assumptions about the failure time distribution. Furthermore, since it is the nonparametric maximum likelihood estimator (NPMLE) of $S$, it is a natural choice among the set of all nonparametric estimators. Most standard scientific and/or statistical software (MATLAB, Mathematica, R, etc.) possess built-functions for calculating the Kaplan-Meier estimate from censored data, making implementation trivial. In situations where the distribution of failure times follows a known parametric distribution, a parametric approach to estimating the marginal survivor function may be chosen instead. 

The Kaplan-Meier estimator may be biased if censoring is informative, i.e., if the failure and censoring times are dependent. In this paper, where we aim to describe a simple extension of NB to time-to-event data, we will operate under the assumption that censoring occurs at random. Formally, we assume that $P(T \geq t, C \geq t) = P(T \geq t) P(C \geq t)$.  The assumption can be slightly weakened if there is a known set of discrete-valued variables $\bY$ such that the failure and censoring times are conditionally independent given $\bY$, in which case one could consider estimating $S(t)$ as $\int_{\bmy} P(T \geq t \given \bY=\bmy) dF_{\bY}(\bmy)$, where $P(T \geq t \given \bY=\bmy)$ can be consistently estimated via Kaplan-Meier within the subset of individuals with $\bY = \bmy$. Random censoring may be a reasonable assumption in EHR datasets where most censoring is induced either by the end of the data capture window or by subjects disenrolling from the healthcare plan, often due to an employer switching plans. Even in cases where random censoring does not strictly hold, the NB approach can be viewed as trading off some bias in estimation of $S(t)$ for additional flexibility in estimation of $S_{\bX}(t)$ via  modeling of the conditional covariate densities $f_j^\geq$ and $f_j^<$.

%

\subsection{Estimation of covariate densities $f_j^\geq$ and $f_j^<$}
\label{sect:parsmooth}

If there are $p$ total covariates, then the Naive Bayes assumption requires estimation of the densities $f_j^\geq(x;t)$ and $f_j^<(x;t)$ for $j=1,\dots,p$ at every $t$. For fixed $t$, one could estimate these univariate conditional densities quite flexibly using kernel density methods, but one is then left with the challenge of combining the resulting density estimates across adjacent values of $t$ to yield relatively smooth functions. In preliminary work, we discovered that the kernel density estimation approach was highly variable and performed poorly, so instead we propose much simpler parametric models yielding density estimates which vary smoothly in $t$.  Specifically, we assume that
\begin{align*}
X_j \given T \geq t \ \sim \ N(\mu_j(t), \sigma^2_j(t)) \ \  &\equiv \ \  f_j^\geq(x;t) = \phi \left( \frac{ x - \mu_j(t)}{ \sigma_j(t) } \right)\\
X_j \given T < t \ \sim \ N(\theta_j(t), \psi^2_j(t)) \ \  &\equiv \ \   f_j^<(x;t) = \phi \left( \frac{ x - \theta_j(t)}{ \psi_j(t) } \right)
\end{align*}
where $\phi$ is the standard Normal PDF. The assumption that changes in the covariate distributions with $t$ are restricted to location-scale shifts within the Normal family is a strong one, but our simulations demonstrate that it need not hold exactly for NB to perform well.

For failure and censoring times $t_k \in \{ t_1, \dots, t_K \}$, consider the following non-parametric estimators of $\mu(t_k), \sigma^2(t_k), \theta(t_k),$ and $\psi^2(t_k)$:
\begin{align}
\hat{\mu}_j(t_k) &= \frac{ \sum_{i=1}^n \indicatorBig{O_i \geq t_k} X_{ij} }{ \sum_{i=1}^n \indicatorBig{O_i \geq t_k} }  \ \ \text{ if } \sum_{i=1}^n \indicatorBig{O_i \geq t_k} \neq 0, \text{ otherwise } \hat \mu_j(t_k) = 0 \label{eq:NPmu}, \\
\hat{\sigma}^2_j(t_k) &= \frac{ \sum_{i=1}^n \indicatorBig{O_i \geq t_k} X^2_{ij} }{ \sum_{i=1}^n \indicatorBig{O_i \geq t_k} }  - \hat\mu^2_j(t_k)  \ \ \text{ if } \sum_{i=1}^n \indicatorBig{O_i \geq t_k} \neq 0, \text{ otherwise } \hat \sigma^2_j(t_k) = 1, \label{eq:NPsig}\\
\hat{\theta}_j(t_k) &= \frac{ \sum_{i=1}^n \delta_i \indicatorBig{O_i < t_k} X_{ij} }{ \sum_{i=1}^n \delta_i \indicatorBig{O_i < t_k} } \ \ \text{ if } \sum_{i=1}^n \delta_i \indicatorBig{O_i < t_k} \neq 0, \text{ otherwise } \hat \theta_j(t_k) = 0, \label{eq:NPtheta}\\
\hat{\psi}^2_j(t_k) &= \frac{ \sum_{i=1}^n \delta_i \indicatorBig{O_i < t_k} X^2_{ij} }{ \sum_{i=1}^n \delta_i \indicatorBig{O_i < t_k} } - \hat\theta^2_j(t_k) \ \ \text{ if } \sum_{i=1}^n \delta_i \indicatorBig{O_i < t_k} \neq 0, \text{ otherwise } \hat \psi^2_j(t_k) = 1. \label{eq:NPpsi}
\end{align}
In \eqref{eq:NPmu} and \eqref{eq:NPsig}, all individuals with observation times $\geq t_k$ contribute to estimation of $\mu$ and $\sigma^2$, since $O_i \equiv \min(T_i, C_i) \geq t_k$ implies that $T_i \geq t_k$. In \eqref{eq:NPtheta} and \eqref{eq:NPpsi}, only observations with observed failures prior to $t_k$ contribute to estimation of $\theta$ and $\psi^2$. The procedure excludes censored observations with $O_i < t_k$, for which it is possible that $T_i \geq t_k$ or $T_i < t_k$. We show in Appendix A that, for event times yielding non-zero denominators, \eqref{eq:NPmu}-\eqref{eq:NPpsi} are consistent estimators of their respective parameters under the following conditions:\\
Condition (1). $T \perp C \given X_j \ \text{for all } j$\\
Condition (2). $C \perp X_j \ \text{for all } j$

Condition (1) is the standard assumption of independence of failure and censoring times, conditional on $X_j$. Condition (2) requires that the censoring time be independent of each $X_j$. This assumption may be plausible in settings (such as EHR data) if censoring is largely administrative in nature, induced by the fact some subjects enrolled near the end of the study period, or occurring due to factors unlikely to be related to failure time (e.g., a subject's employer switching health plans). Together, Conditions (1) and (2) imply that $T \perp C$, i.e., failure and censoring times are marginally independent.


\subsubsection{Smoothing of covariate densities}

The estimators \eqref{eq:NPmu}-\eqref{eq:NPsig} may fluctuate substantially between adjacent failure times as outlying covariate values are either excluded from the risk set or acquire more influence as less extreme values are removed. Furthermore, these estimators leave undefined values of the densities for $t_k < t < t_{k+1}$. We therefore propose to apply nonparametric smoothing techniques to estimate the functions $\mu_j(t), \sigma_j^2(t), \theta_j(t),$ and $\psi^2_j(t)$ . One popular smoothing technique is \textit{loess} \cite{Cleveland1988}, which combines local linear fits across a moving window. Since the precision of the estimates of $\hat{\mu}_j(t_k)$ and $\hat{\sigma}_j^2(t_k)$ decreases as $t_k$ increases and the precision of $\hat \theta_j(t_k)$ and $\hat \psi^2_j(t_k)$ decreases as $t_k$ decreases, we propose to fit the loess curves to observations weighted according to their estimated precision. For the mean curve $\mu_j(t)$, we fit a loess curve to the weighted data
\[
w_1 \hat{\mu}_j(t_1), w_2 \hat{\mu}_j(t_2), \dots, w_K \hat{\mu}_j(t_K)
\]
where $w_k = \frac{1}{s^2_k/n_k}$ with $s^2_k$ the empirical variance of $X_j$ observations and $n_k = \sum_{i=1}^n \indicatorBig{O_i \geq t_k}$. For $\hat{\sigma}^2_j$, the weights are given by $w_k = \frac{1}{2 s^4_k / (n_k - 1)}$, where $s^4 = (s^2)^2$. $\hat \theta_j$ and $\hat \psi^2_j$ are weighted by corresponding formulae, with $n_k = \sum_{i=1}^n \delta_i \indicatorBig{O_i < t_k}$ and $s^2_k$ the empirical variance of $X_j$ among subjects with failure times prior to $t_k$.

The final algorithm for estimating the conditional survivor function $P( T \geq t \given \bX)$ is therefore as follows:\\

Given a set of failure times $\scriptT = \{t_1, \dots, t_K\}$:
\begin{enumerate}
\item For every covariate $X_j$, do the following:
	\begin{enumerate}
	\item Evaluate $\hat{\mu}_j(t), \hat \sigma^2_j(t), \hat \theta_j(t)$, and $\hat \psi^2_j(t)$ for $t \in \scriptT$.
	\item Fit a weighted loess smoother to plots of the resulting estimates versus $t$, for $t \in \scriptT$. Let the resulting estimated nonparametric regression functions be known as $\hat \mu^*_j, \hat \sigma^{2,*}_j, \hat \theta^*_j,$ and $\hat \psi^{2,*}_j$.	
	\item Plug in $\hat \mu^*_j$ and $\hat \sigma^{2,*}_j$ to estimate $f_j^\geq(t)$ and $\hat \theta^*_j$ and $\hat \psi^{2,*}_j$ to estimate $f_j^<(t)$ assuming that $X_j \given T \geq t \sim N( \hat \mu^*_j(t), \hat \sigma^{2,*}_j(t) )$ and $X_j \given T < t \sim N( \hat \theta^*_j(t), \hat \psi^{2,*}_j(t))$.
	\item Estimate the marginal survivor function $S(t)$ using the Kaplan-Meier estimator.
	\end{enumerate}
\item Estimate the survivor function $S_{\bX}(t)$ by plugging in estimates from steps 3 and 4 into \eqref{eq:naivebayes}. If the plugin estimates yield an estimated probability greater than $1$, report a predicted probability of $1$.
\end{enumerate}

Software implementing this extended Naive Bayes technique is available on the website of the first author (Wolfson). In the following sections, we illustrate the application of our method to simulated data, compare performance with the Cox proportional hazards model, and use the NB to build a cardiovascular risk prediction model using electronic health record data.

\section{Simulations} \label{sect:sims}

One theoretical advantage of the proposed Bayesian network approach is that it does not depend on correct specification of a regression model for the hazard function (as in the Cox model). Hence, NB might be better suited to modeling data which do not follow a proportional hazards model, or for which the functional relationship between covariates and the hazard is misspecified. In this section, we compare the performance of our Naive Bayes method with the Cox proportional hazards model (CPH) when 1) failure times follow a proportional hazards model, 2) failures times follow a log-logistic accelerated failure time (AFT) model such that the hazards are non-proportional, and 3) the manner in which covariates influence the failure time is misspecified in the regression model. In all cases, censoring times were generated as $\min(10, Unif(0,20))$, thereby guaranteeing the conditional and unconditional independence of censoring and failure times. Observed event times $O$ and censoring indicators $\delta$ were defined in the usual way: $O = \min(T,C)$ and $\delta = \indicatorBig{T \leq C}$. We consider predicting the probability of surviving beyond 7 years, $S_{\bX}(7) = P(T \geq 7 \given \bX)$, though we emphasize that our approach yields estimates of the entire conditional survivor function so that  $S_{\bX}(t)$ could be predicted for any $t$ with a trivial amount of additional computation. 

\subsection{Performance metrics}

In our simulation, we sought to evaluate two characteristics of the NB and CPH techniques: model calibration, the degree to which predicted and true event probabilities coincide; and model discrimination, the degree to which the model can identify low- and high-risk individuals. In all cases, calibration and discrimination were assessed on separately simulated validation datasets of equal size. All tables are based on 1000 simulations.

\subsubsection{Model calibration.} To assess model calibration, bias and mean squared error (MSE) of the predicted probabilities were computed with respect to the true event probabilities calculated from the various simulation models. We report ``conditional'' bias and MSE within specific population subgroups, as well as ``marginal'' measures which average across the entire simulated population.

\subsubsection{Model discrimination.} To compare the ability of the CPH and NB models to predict events (in this case, a failure prior to time $t=7$), we used the net reclassification improvement (NRI) \cite{Pencina2008}. The NRI, also referred to as the net reclassification index, compares the number of ``wins'' for each model among discordant predictions. We defined risk quartiles according to the true event probabilities; risk predictions for an individual were considered discordant between the NB and CPH models if the predictions fell in different quartiles. The NRI for comparing the NB and CPH models is defined by
\begin{equation}
NRI = \frac{E_{nbn}^{\uparrow} - E_{cph}^{\uparrow}}{n_E} + \frac{ \bar{E}_{nbn}^{\downarrow} - \bar{E}_{cph}^{\downarrow}}{ n_{\bar{E}}}
\label{eq:NRI}
\end{equation}
$E_{nbn}^{\uparrow}$ are the number of individuals who experienced events prior to 7 years and were placed in a higher risk category by the NB than the CPH algorithm (the opposite change in risk categorization yields $E_{cph}^{\uparrow}$). $\bar{E}_{nph}^{\downarrow}$ and $\bar{E}_{cph}^{\downarrow}$ count the number of individuals who did not experience an event and were ``down-classified'' by NB and CPH, respectively. $n_E$ and $n_{\bar E}$ are the total number of events and non-events, respectively. Event indicators were defined by whether the simulated failure time $T$ was less than 7, regardless of whether the failure was observed or not. A positive NRI means better reclassification performance for the NB model, while a negative NRI favors the CPH model. While NRIs can theoretically range from -1 to 1, in the risk prediction setting they do not typically exceed the range [-0.25, 0.25].  For example, \cite{Cook2009} calculated the effects of omitting various risk factors from the Reynolds Risk Score model for prediction of 10-year cardiovascular risk. The estimated NRIs ranged from -0.195 (omitting age) to -0.032 (omitting total cholesterol or parental history of MI), and all were statistically significant at the 0.05 level. In addition to the NRI, the tables of simulation results also display the reclassification improvement rates for events and non-events (i.e. the summands in \eqref{eq:NRI}) separately. In our simulations, the true event times of each individual in the test set were used to calculate the NRI. In the real data analyzed in Section \ref{sect:application}, the test data are subject to censoring, hence we use a modified version of the NRI due to \cite{Pencina2011} which accounts for censoring.

\subsection{Proportional hazards}

Table \ref{tab:PH} summarizes the performance of the CPH and NB models when data are generated from a proportional hazards model using the method suggested in \cite{Bender2005}. Covariates $\bX_c = (X_1, X_2, \dots, X_p)$ were generated from a $p$-variate Normal distribution with mean zero, variance one, and exchangeable correlation with pairwise correlation $\rho$. Different values of $\rho$ were considered in the various simulation settings. The final design matrix was defined by $\bX = [ \boldmath{1} \ \bX_c ]$, where $\mathbf{1}$ is an $n$-vector of ones. Because each covariate was marginally standard normal, no covariate scaling was performed before applying the computational methods. Given observed covariates $\bx_i$ for subject $i$, a failure time was generated as 
\[
T_i = -\left( \frac{ \log(U_i) }{ \lambda \exp(\bbeta' \bx_i) } \right)^\nu
\]
where $U_i \sim Unif(0,1)$, which corresponds to generating failure times from a Weibull distribution with scale parameter $\lambda(\bx_i) = \lambda \exp(\bbeta'\bx_i)$, so that the hazard for subject $i$ is proportional to the baseline hazard with proportionality factor $\exp(\bbeta'\bx_i)$. For this set of simulations, we set $\lambda = 0.01$, $\nu=2$, and 
\[
\bbeta = (\beta_0, 0.5, 0,0,0,0)
\]
$\beta_0$ was set to $0, -1$, and $-2$ across simulations, yielding censoring rates of 55\%, 78\%, and 90\%, and event rates of 40\%, 18\%, and 7\% respectively. 

\begin{table}[ht]
\begin{center}
\begin{tabular}{ccc|ccccccc}
  \hline
$n$ & $\beta_0$ & $\rho$ & Bias(CPH) & Bias(NB) & MSE(CPH) & MSE(NB) & RI(E) & RI(NE) & NRI \\
  \hline
1000 & 0 & 0.0 & 0.020 & -0.002 & 0.160 & 0.280 & 0.150 & -0.160 & -0.006 \\
  1000 & 0 & 0.7 & 0.020 & -0.020 & 0.160 & 4.300 & 0.081 & -0.041 & 0.040 \\
  1000 & -1 & 0.0 & 0.006 & -0.000 & 0.088 & 0.210 & 0.048 & -0.069 & -0.021 \\
  1000 & -1 & 0.7 & 0.004 & -0.057 & 0.087 & 3.500 & 0.000 & 0.045 & 0.045 \\
  1000 & -2 & 0.0 & 0.001 & 0.000 & 0.042 & 0.110 & -0.032 & -0.004 & -0.037 \\
  1000 & -2 & 0.7 & 0.002 & -0.043 & 0.041 & 1.900 & 0.010 & 0.035 & 0.045 \\
\hline
  5000 & 0 & 0.0 & 0.020 & 0.000 & 0.067 & 0.055 & 0.150 & -0.150 & 0.001 \\
  5000 & 0 & 0.7 & 0.020 & -0.019 & 0.068 & 4.300 & 0.078 & -0.041 & 0.036 \\
  5000 & -1 & 0.0 & 0.005 & 0.000 & 0.020 & 0.042 & 0.052 & -0.059 & -0.007 \\
  5000 & -1 & 0.7 & 0.005 & -0.056 & 0.020 & 3.500 & -0.021 & 0.050 & 0.030 \\
  5000 & -2 & 0.0 & 0.001 & 0.000 & 0.009 & 0.022 & 0.002 & -0.016 & -0.014 \\
  5000 & -2 & 0.7 & 0.001 & -0.046 & 0.008 & 2.000 & -0.061 & 0.095 & 0.033 \\
   \hline
\end{tabular}
\end{center}
\caption{Performance of CPH and NB models on data simulated from Weibull (proportional hazards) model. Mean squared errors are computed over the entire validation set and multiplied by 100. RI(E) and RI(NE) are the reclassification improvement rates for events and non-events, and their sum yields the NRI. Positive reclassification improvement scores favor NB, and negative scores favor CPH. \label{tab:PH}}
\end{table}

\begin{figure}
\centering
	\begin{subfigure}{\textwidth}
	\centering
	\includegraphics[width=0.8\textwidth]{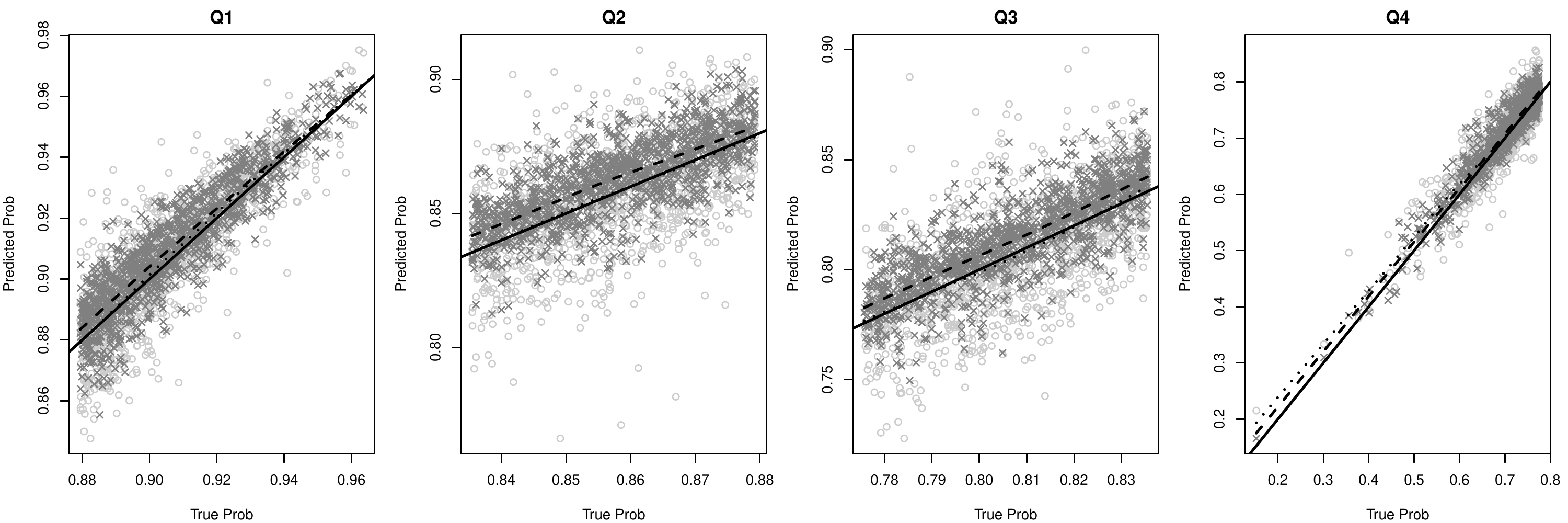}
	\caption{$\rho = 0$}
	\label{fig:PHcalib1}
	\end{subfigure}

	\begin{subfigure}{\textwidth}
	\centering
	\includegraphics[width=0.8\textwidth]{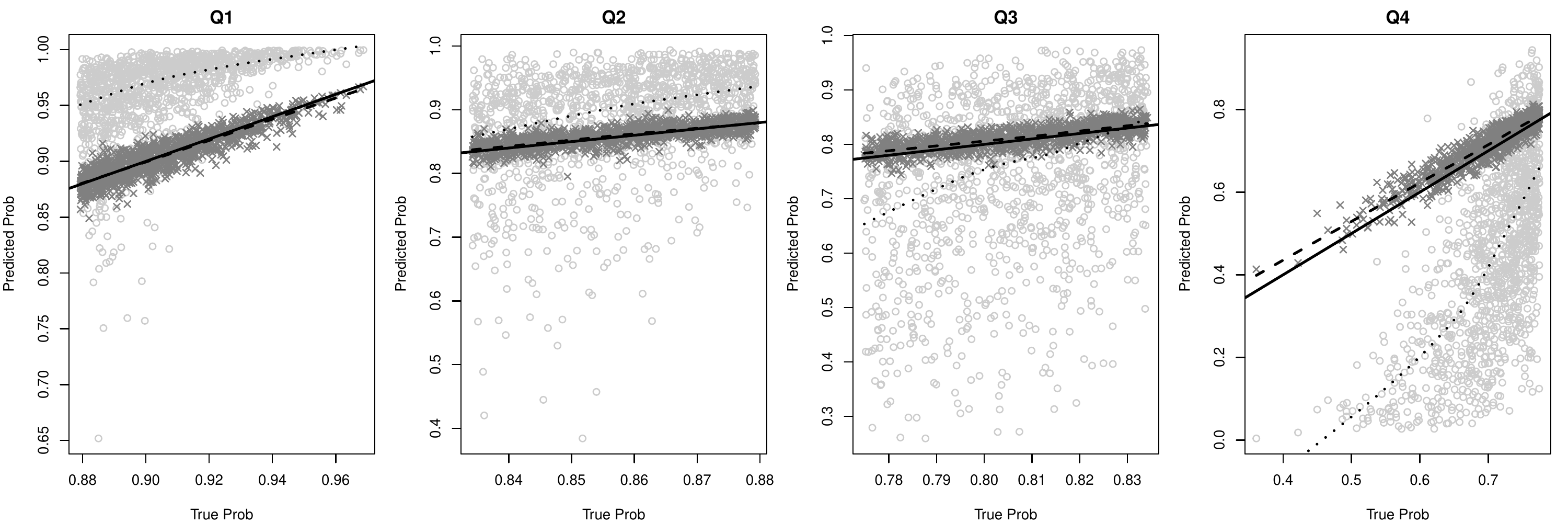}
	\caption{$\rho=0.7$}
	\label{fig:PHcalib2}
	\end{subfigure}
\caption{Predicted vs. true probabilities across quantiles of $X_1$ (coefficient $\beta_1 = 0.5$) for 5000 points sampled from 10 simulation runs under the settings where $\rho=0$ and $\rho=0.7$. $\beta_0$ is fixed at -1. Dark gray points are predictions from CPH, light gray from NB. Dashed line is a lowess smoother applied to CPH predictions, dotted line a smoother applied to NB predictions.}
\end{figure}

Figures \ref{fig:PHcalib1} and \ref{fig:PHcalib2} plot predicted vs. true probabilities across quantiles of $X_1$ (coefficient $\beta_1 = 0.5$) for 5000 points sampled from 10 simulation runs under the settings where $\rho=0$ and $\rho=0.7$, respectively ($\beta_0$ is fixed at -1). 

When covariates are independent ($\rho =0$), NB and CPH have similar calibration and discrimination performance. When covariates are strongly correlated, NB has an anti-conservative bias in which large survival probabilities are overestimated and small survival probabilities are underestimated (see, e.g., panels 1 and 4 of Figure \ref{fig:PHcalib2}). Interestingly, this bias does not degrade the discrimination ability of NB as measured by the NRI. In fact, the NRI favors NB over CPH more strongly in cases where bias is larger. This finding is not entirely surprising; it has been shown that NB tends to drive predicted probabilities towards either 0 or 1 \cite{Zadrozny2001} while retaining an approximately correct ranking of observations with respect to their risk.


\subsection{Nonproportional hazards}

Table \ref{tab:NPH} compares the performance of the NB with the CPH model when data follow a log-logistic accelerated failure time (AFT) model. Covariates were generated as described above. Given observed covariate $\bx_i$, failure times were generated from a Log-logistic distribution with shape parameter $\exp(\bbeta' \bx_i)$ and scale parameter $\phi=20$.  Censoring times were generated as $\min(10,Unif(0,20))$. $\bbeta$ was set as
\[
\bbeta = (\beta_0, 0.5, 0.1, -0.1, 0, 0)
\]
$\beta_0$ was set to $1, 0$, and $-1$ across simulations, yielding censoring rates of 60\%, 74\%, and 90\%, and event rates of 40\%, 25\%, and 8\% respectively.

\begin{table}
\begin{center}
\begin{tabular}{ccc|ccccccc}
  \hline
$n$ & $\beta_0$ & $\rho$ & Bias(CPH) & Bias(NB) & MSE(CPH) & MSE(NB) & RI(E) & RI(NE) & NRI \\
  \hline
  1000 & 1 & 0.0 & 0.002 & 0.000 & 0.100 & 0.140 & 0.024 & 0.180 & 0.210 \\
  1000 & 1 & 0.7 & 0.004 & -0.099 & 0.097 & 5.800 & 0.018 & 0.270 & 0.290 \\
  1000 & 0 & 0.0 & 0.007 & 0.001 & 0.220 & 0.260 & 0.170 & -0.130 & 0.042 \\
  1000 & 0 & 0.7 & 0.007 & -0.058 & 0.200 & 4.100 & 0.160 & -0.091 & 0.064 \\
1000 & -1 & 0.0 & 0.005 & -0.001 & 0.230 & 0.310 & 0.150 & -0.160 & -0.011 \\
  1000 & -1 & 0.7 & 0.005 & -0.012 & 0.210 & 1.800 & 0.180 & -0.160 & 0.015 \\
 \hline
  5000 & 1 & 0.0 & 0.003 & 0.001 & 0.056 & 0.037 & 0.053 & 0.180 & 0.230 \\
  5000 & 1 & 0.7 & 0.002 & -0.100 & 0.052 & 6.000 & -0.005 & 0.310 & 0.300 \\
  5000 & 0 & 0.0 & 0.005 & 0.001 & 0.110 & 0.068 & 0.190 & -0.140 & 0.050 \\
  5000 & 0 & 0.7 & 0.005 & -0.059 & 0.110 & 4.000 & 0.140 & -0.087 & 0.057 \\
  5000 & -1 & 0.0 & 0.004 & 0.000 & 0.100 & 0.071 & 0.200 & -0.200 & -0.003 \\
  5000 & -1 & 0.7 & 0.004 & -0.012 & 0.096 & 1.500 & 0.200 & -0.190 & 0.009 \\
   \hline
\end{tabular}
\end{center}
\caption{Performance of CPH and NB models on data simulated from Log-logistic (non-proportional hazards) model. Mean squared errors are multiplied by 100. RI(E) and RI(NE) are the reclassification improvement rates for events and non-events, and their sum yields the NRI. Positive reclassification improvement scores favor NB, and negative scores favor CPH.  \label{tab:NPH}}
\end{table}

As in the proportional hazards case, bias was small for both methods when covariates were independent, and was higher for NB when covariates were correlated. NB performed better than CPH in terms of risk reclassfication, with the advantage being particularly dramatic ($> 20\%$ net reclassification improvement) when the underlying event rate was higher. In general, the NB yielded better risk classification for subjects who experienced events, while CPH performed better for subjects without events.

\subsection{Regression model misspecification}
\label{sect:misspecification}

Table \ref{tab:OmCov} summarizes the performance of NB and CPH models when data are generated from a Log-logistic model and the CPH linear predictor is misspecified. Our simulation is motivated by our applied data example, where the effects of age and systolic blood pressure (SBP) on the probability of experiencing a nonfatal cardiovascular event appear to be non-linear. The simulation model encodes a scenario where subjects at middle ages are at highest risk, while those at low and high ages are at lower risk.

\begin{figure}
\centering
\includegraphics[width=0.8\textwidth]{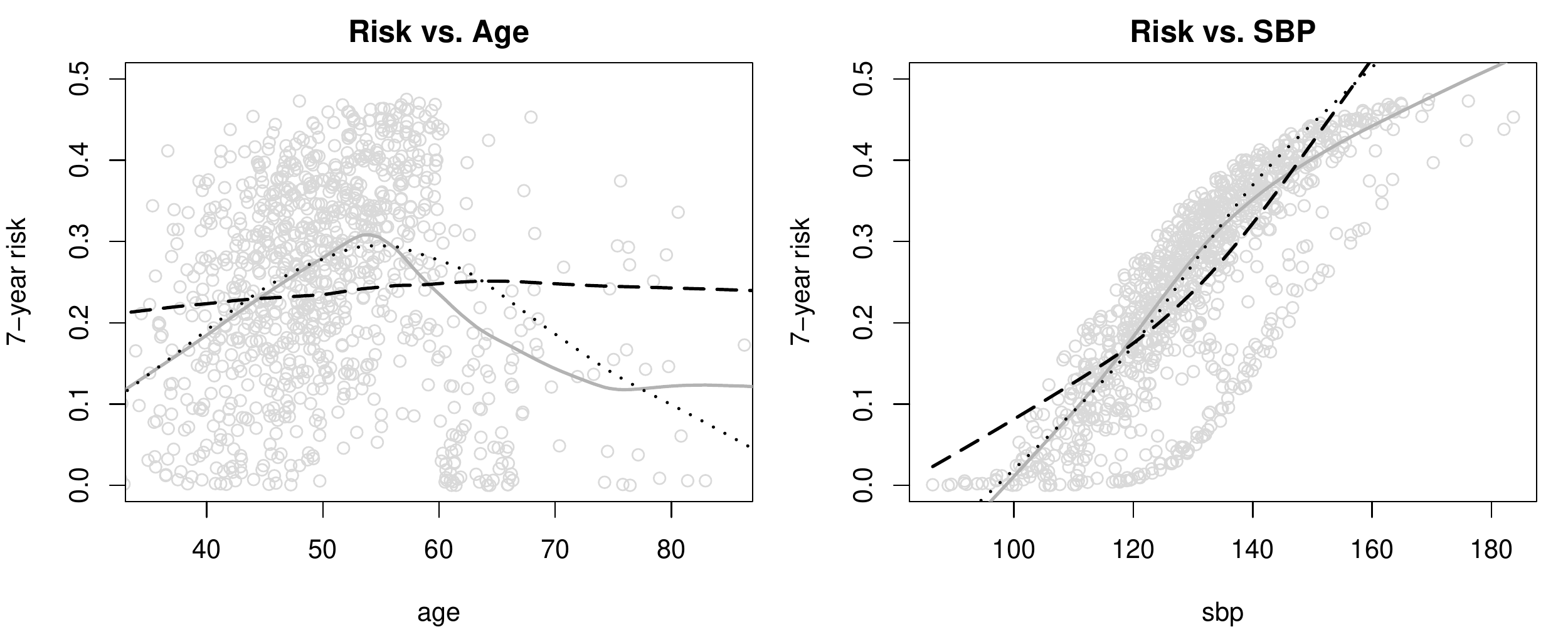}
\caption{True and predicted risks of experiencing an event based on a Log-logistic model with shape parameter defined by \eqref{eq:regmodel}, plotted versus age (left panel) and SBP (right panel). On each plot, the points show (for a typical simulation run) true simulated risks and the solid line is a lowess scatterplot smooth showing the trend in risk across age and SBP respectively. The dashed lines in each panel show the trend in risk predictions obtained from the Cox Proportional Hazards model incorporating linear main effects for age and SBP (first row of Table \ref{tab:OmCov}), while the dotted lines show the trend in predictions by Naive Bayes. \label{fig:OmCov}}
\end{figure}

For this set of simulations, the total sample size was fixed at $n=5000$. We started by generating an `age' covariate from a Log-logistic distribution with shape parameter 10, scale parameter 50, and rate parameter 0.4; this yielded a distribution of ages with a median of 50 and an interquartile range of 10. A small fraction ($<0.1\%$) of simulated ages were $> 100$; these ages were set to 100. Given age$=a$, systolic blood pressure values were simulated from a Normal distribution with mean $130+sign(a-50)\sqrt{|a-50|}$ and standard deviation $15$. Based on the simulated ages and SBPs, failure times were simulated from a Log-logistic distribution with scale parameter $\phi=20$ and shape parameter 
\begin{equation}
\lambda(age,sbp) = \exp \left[ -0.2 \cdot age -0.6 \cdot SBP +0.2  \cdot (age>60) -0.2 \cdot age \times SBP + 0.4 \cdot age \times (age>60) \right]
\label{eq:regmodel}
\end{equation}
where $(age>60)$ is a binary indicator, $age \times sbp$ is an age-by-SBP interaction, and $age \times (age>60)$ is an interaction between continuous age and binary indicator of age $>60$. All columns of the design matrix implied by this linear predictor were standardized to have mean zero and variance one before estimation was performed, so that the coefficients represent standardized effect sizes for each covariate. The left panel of Figure \ref{fig:OmCov} (plot labeled `True') shows a set of survival probabilities (i.e. probabilities of having a failure time $\geq 7$ years) simulated from this model and plotted vs. age. 

We consider the performance of the NB and CPH models for different specifications of the linear predictor in the Cox model. There is no explicit regression model associated with the NB, and hence it is provided with (standardized) vectors of ages and SBPs. We consider varying degrees of misspecification of the regression model, from a model incorporating only linear terms for age and SBP (omitting the (age $> 60$) indicator and higher-order interactions) to the full model including all the covariate terms from \eqref{eq:regmodel}.

\begin{table}[ht]
\begin{center}
\begin{tabular}{l|ccccccc}
 \hline
CPH regression terms & Bias(CPH) & Bias(NB) & MSE(CPH) & MSE(NB) & RI(E) & RI(NE) & NRI \\
\hline
Age, SBP & 0.008 & 0.000 & 0.623 & 0.419 & 0.175 & -0.061 & 0.113 \\
 log(Age), SBP & 0.008 & 0.000 & 0.634 & 0.421 & 0.179 & -0.060 & 0.118 \\
 Age, SBP, Age x SBP & 0.008 & 0.000 & 0.615 & 0.422 & 0.166 & -0.059 & 0.107 \\
  Age, SBP, (Age$^2$) & 0.009 & 0.001 & 0.498 & 0.421 & 0.127 & -0.067 & 0.060 \\
  Age, SBP, (Age$>$60) & 0.011 & 0.000 & 0.268 & 0.422 & 0.078 & -0.107 & -0.029 \\
All & 0.011 & 0.001 & 0.272 & 0.422 & 0.076 & -0.109 & -0.032 \\
   \hline
\end{tabular}
\end{center}
\caption{Performance of CPH and NB models for data simulated from a Log-logistic regression model including the variables from \eqref{eq:regmodel}. The total sample size for each simulation was $n=5000$. The first column of the table specifies which terms are included in the CPH regression model; the NB is provided with age and SBP values only. The final row shows the results when the CPH model includes all relevant covariates. RI(E) and RI(NE) are the reclassification improvement rates for events and non-events, respectively. Mean squared errors are multiplied by 100. \label{tab:OmCov}}
\end{table}

Table \ref{tab:OmCov} and Figure \ref{fig:OmCov} illustrate how the additional flexibility of the NB model allows more accurate predictions to be obtained when relevant covariates are omitted from the model. The NB model has smaller MSE and positive NRI when compared to a CPH model which includes only main effects for age (or log(Age)) and SBP. Figure \ref{fig:OmCov} displays predicted survival probabilities from a typical simulation run using the NB and the main effects only CPH model. The overlaid smoothers show clearly that the NB captures the non-linear effects of age and SBP observed in the true probabilities (left panel) better than CPH.  Returning to Table \ref{tab:OmCov}, adding an age $\times$ SBP interaction term does not markedly improve the performance of CPH. Incorporating a quadratic effect for age, a standard strategy when faced with apparent non-linearity in the effect of a covariate, yields some improvement in CPH but the NB retains an advantage in MSE and NRI. When the (Age$>60$) indicator term is included, the CPH model outperforms NB both in terms of MSE and NRI.
 
\section{Data application}
\label{sect:application}

We illustrate our Naive Bayes method by applying it to the task of predicting the risk of cardiovascular events from longitudinal electronic health record data. The dataset includes medical information on a cohort of patient-members age 18 years and older from a large Midwest integrated healthcare system. Data were available on members who attended a primary care clinic between 2002 and 2009, and who were continuously enrolled for at least twelve consecutive months during that period. 

We consider the task of predicting the individual risk of experiencing a cardiovascular event, including death where cardiovascular disease was listed as the primary cause; procedures such as thromboectomy, stent placement or coronary artery bypass graft; and diagnoses such as stroke or mycocardial infraction. For both event definitions, failure and censoring times of subjects were measured from an index visit, corresponding to the first clinic visit following 12 months of continuous enrollment in the healthcare system. Only first events were recorded; subjects were censored if they disenrolled prior to experiencing an event or were event-free when the dataset was locked and prepared for analysis in 2009. Those whose primary cause of death was not cardiovascular disease were coded as censored. In this data application example, we did not adjust either the CPH or NB methods for competing risks. 

Information available for predicting the risk of cardiovascular events included age, blood pressure, and LDL cholesterol. We restricted our analysis to subjects who had values for these risk factors recorded prior to the index visit; these factors were treated as baseline predictors. The one exception was cholesterol, which was often missing for subjects under the age of 50. Due to low rates of dyslipidemia in this group, subjects without a measured LDL cholesterol within the first year of enrollment were assigned a normal LDL cholesterol value of 120 mg/dl. With these restrictions, the analysis cohort consisted of 274,964 patients with a mean follow up of 4.8 years. The cohort was randomly split into a training dataset consisting of 75\% of the observations (206,223 subjects), and a test dataset consisting of the remaining 25\% (68,741 subjects). After standardizing the covariate values, we applied both the CPH and NB methods to the training data and obtained predictions for the test data. The CPH model regression equation included linear terms for each of the three risk factors (age, blood pressure, and LDL cholesterol), consistent with existing CPH-based techniques such as the Framingham risk equations.  

\begin{figure}
\centering
\includegraphics[width=3.5in]{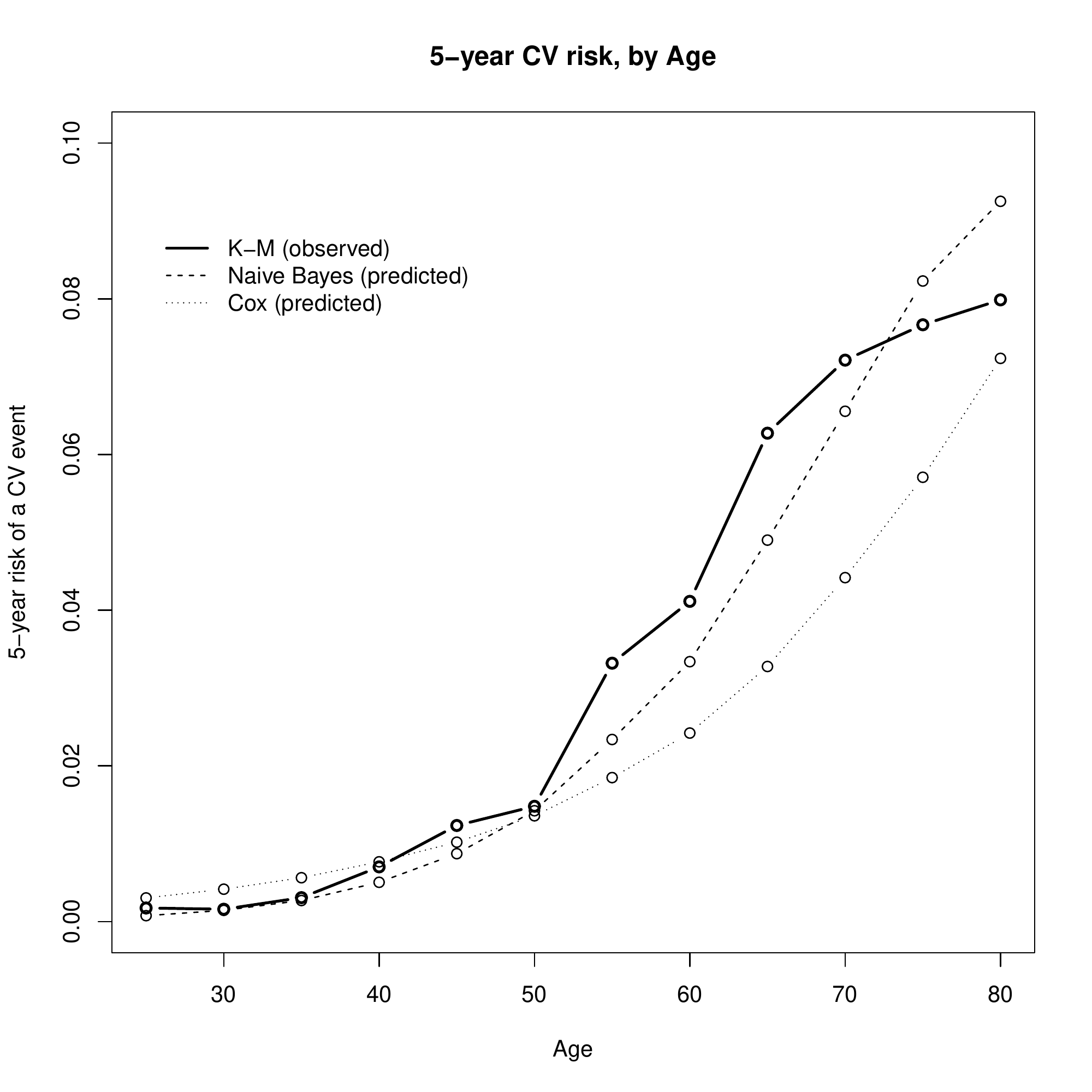}
\caption{Observed and predicted 5-year cardiovascular event rates vs. age. \label{fig:HP}}
\end{figure}

Figure \ref{fig:HP} shows five-year cardiovascular risk predicted by NB and CPH, by age, for the two event definitions. Overlaid on the two plots is the observed event rate for five-year age ranges, computed within each range via Kaplan-Meier. The figure illustrates how the flexibility of the NB method allows it to capture a marked uptick in event rate after the age of 50; in contrast,  the linear predictor of the Cox model restricts the set of age-risk relationships that can be well approximated.

\subsection{Reclassification with censored observations}

As noted earlier, the Net Reclassification Improvement statistic cannot be applied directly to data where the outcome status of some subjects is unknown. In our setting, omitting subjects with less than five years of follow-up (or treating them as non-events) will result in biased estimates of the NRI.  To evaluate risk reclassification on our test data which are subject to censoring, we use a ``censoring-adjusted'' NRI (cNRI) which takes the form

\begin{equation}
cNRI = \frac{E_{nbn}^{*,\uparrow} - E_{cph}^{*,\uparrow}}{n^*_E} + \frac{ \bar{E}_{nbn}^{*,\downarrow} - \bar{E}_{cph}^{*,\downarrow}}{ n^*_{\bar{E}}}
\label{eq:cNRI}
\end{equation}

Full details appear in \cite{Pencina2011}; briefly, the expected number of events overall and among up-classified and down-classified individuals is computed from event probabilities estimated via Kaplan-Meier. Uncertainty intervals are obtained by computing the cNRIs on the test set for predictions from NB and Cox models fitted to 500 bootstrap resamples of the training data.


Using the test data ($N=68,741$), we computed the cNRI comparing the two models.  Following the recommendations of \cite{Pepe2011}, we considered clinically relevant risk categories for five-year risk of cardiovascular events: 0-5\%, 5-10\%, and $> 10\%$. Table \ref{tab:appNRI} summarizes the reclassification improvements and net reclassification improvements for subjects with and without events. The NRI across all subjects was positive (NRI = 0.10, 95\% CI 0.084 to 0.120), indicating that NB had significantly better reclassification performance than the Cox model. The reclassification improvements behave similarly to our simulations, with Naive Bayes outperforming the Cox model in terms of predicting the occurrence of events among those who actually experienced them, and the Cox model apparently more accurate among subjects who did not experience events.

\begin{table}
\centering
\begin{tabular}{cc|ccc}
 $N$ (test set) & \# of events & cRI (events) & cRI (non-events) & cNRI\\
 \hline
68,741 & 1,382 & $\mathbf{0.14 \pm 0.030}$ & $\mathit{-0.04 \pm 0.006}$ & $\mathbf{0.10 \pm 0.021}$ \\
\end{tabular}
\caption{Censoring-adjusted Reclassification Improvement and Net Reclassfication Improvement (cRIs and cNRIs) comparing Naive Bayes and Cox models for predicting cardiovascular events from 5295 patients' electronic health record data. Positive cRIs and cNRIs favor Naive Bayes, and negative cNRIs favor the Cox model. $A \pm B$ denotes an estimated cRI/cNRI of $A$ with 95\% Wald confidence interval $(A - B, A+B)$. Reclassification improvement scores that are significantly different than zero are indicated in \textbf{bold} (favoring Naive Bayes) and \textit{italics} (favoring Cox) \label{tab:appNRI}}
\end{table}

\section{Discussion}
We have presented an extension of the Naive Bayes technique to accommodate time-to-event data subject to censoring, and illustrated its application to predicting cardiovascular risk using electronic health record data. Our method, like many machine learning techniques, is `model-agnostic' and hence can capture complex relationships between demographics, biomarkers, and clinical risk without requiring \textit{a priori} specification of a regression model. While it is true that a more complex Cox regression model or a set of stratified Kaplan-Meier estimates might capture observed non-linearities in risk, considering multiple candidate regression models or covariate strata increases the risk of overfitting, even when an independent test set is used. In contrast, our relatively simple Naive Bayes approach captures non-linear effects without requiring any model tuning, scoring, or comparisons. The NB is straightforward to implement and runs quickly; in our simulation studies, unoptimized R code for implementing NB had similar running time to the \texttt{coxph} command from the \texttt{survival} package. In our data application, we successfully applied the NB method to a dataset with $>20,000$ observations, where the total running time was several minutes using a simple implementation in MATLAB. The assumption of covariate independence allows our NB algorithm to be easily parallelized, which is not true of procedures for maximizing the Cox partial likelihood.

The proposed NB method assumes that covariates are independent and normally distributed conditionally on both $T \geq t$ and $T < t$ for all $t$. This is unlikely to hold in practice, though the NB has been shown to be effective in other settings where distributional assumptions regarding the covariates are not satisified. Indeed, while most covariates in our simulation study were generated from marginal normal distributions, no constraint was placed on their conditional distributions. Regarding the issue of calibration, we found that predictions from NB were mostly unbiased and often (but not always) had higher MSE than those from the Cox model. The respectable calibration of the NB predictions may be due to the smoothing of conditional covariate means and variances across time which may attenuate and stabilize probability estimates. The overall effect may be similar to the smoothing and binning approaches described in \cite{Zadrozny2001} for improving calibration of NB classifiers. Perhaps unsurprisingly given its origins, NB outperformed the Cox model in terms of Net Reclassification Improvement, a metric which evaluates how successfully models classify subjects into clinically relevant risk groups. In clinical practice, the ability to discriminate between low- and high-risk patients may be more important than obtaining a precise risk estimate, which may recommend the use of Naive Bayes in settings such as the cardiovascular risk prediction task we consider here.

The covariate independence assumption can be relaxed by modeling covariates as arising from a multivariate normal distribution, but it is unclear how to extend this approach to our setting where parameter values are smoothed over time. Specifically, the key challenge is that covariance matrices must be positive definite, a constraint which is difficult to incorporate into a smoothing procedure. The assumption of conditional normal distributions (given $T \leq t$ and $T > t$) is also somewhat limiting in that it restricts NB to cases where covariates are continuous and have distributions which can be approximated by a normal. In practice, separate NB models could be fit at each level of categorical risk factors (eg. smokers and non-smokers), but this could be time-consuming for factors with more than two levels and inefficient for ordered factors (eg. tumor staging). Furthermore, many risk factors of clinical disease have skewed distributions either marginally or among subjects with survival times beyond a certain threshold. In ongoing work, we are developing techniques which accommodate discrete and skewed covariates in the NB framework.

\section*{Acknowledgments}

This work was partially supported by NHLBI grant \#R01HL102144-01 and AHRQ grant \#R21HS017622-01.

\bibliographystyle{plainnat}
\bibliography{library}

\section*{Appendix A}

\noindent \textit{Theorem.} \\

Let $t_k$ be an event time for which $\sum_{i=1}^n \delta_i \indicatorBig{O_i < t_k} > 0$ and $\sum_{i=1}^n  \indicatorBig{O_i \geq t_k} > 0$. Then
\begin{align}
\hat{\theta}_j(t_k) &= \frac{ \sum_{i=1}^n \delta_i \indicatorBig{O_i < t_k} X_{ij} }{ \sum_{i=1}^n \delta_i \indicatorBig{O_i < t_k} }\\
\hat{\psi}^2_j(t_k) &= \frac{ \sum_{i=1}^n \delta_i \indicatorBig{O_i < t_k} X^2_{ij} }{ \sum_{i=1}^n \delta_i \indicatorBig{O_i < t_k} } - \hat\theta^2_j(t_k)
\end{align}
are consistent for $\theta_j(t_k) \equiv E(X_j \given T < t_k)$ and $\psi^2_j(t_k) \equiv Var(X_j \given T < t_k)$, and
\begin{align}
\hat{\mu}_j(t_k) &= \frac{ \sum_{i=1}^n \indicatorBig{O_i \geq t_k} X_{ij} }{ \sum_{i=1}^n \indicatorBig{O_i \geq t_k} } \\
\hat{\sigma}^2_j(t_k) &= \frac{ \sum_{i=1}^n \indicatorBig{O_i \geq t_k} X^2_{ij} }{ \sum_{i=1}^n \indicatorBig{O_i \geq t_k} }  - \hat\mu^2_j(t_k)
\end{align}
are consistent for $\mu_j(t_k) \equiv E(X_j \given T \geq t_k)$, $\sigma^2_j(t_k) \equiv Var(X_j \given T \geq t_k)$, provided that the following conditions hold:\\
Condition (1). $T \perp C \given X_j \ \text{for all } j$\\
Condition (2). $C \perp X_j \ \text{for all } j$.

\noindent \textit{Proof.}\\

We begin by showing that $\hat{\mu}_j(t) \convto_p E(X_j \given O \geq t)$, where we have replaced $t_k$ by $t$ for clarity. By writing
\begin{equation}
\hat{\mu}_j(t) = \frac{ \frac{1}{n} \sum_{i=1}^n \indicatorBig{O_i \geq t} X_{ij} }{ \frac{1}{n} \sum_{i=1}^n \indicatorBig{O_i \geq t} }
\end{equation}
it is clear that the numerator and denominator converge to $E(\indicatorBig{O \geq t} X_j)$ and $E(\indicatorBig{O \geq t})$ respectively. Let $Y = \indicatorBig{O \geq t}$; then the expression in the denominator is $P(Y=1)$ (which is $> 0$ since $t \leq t_K$) and the numerator can be written as
\begin{align}
E(YX_j) &= E(E(YX_j|Y))\\
	&= E(YX_j|Y=0)P(Y=0) + E(YX_j|Y=1)P(Y=1)\\
	&= 0 + E(X_j|Y=1)P(Y=1)
\end{align}
Converting back to our original notation, we conclude that
\begin{align}
\hat{\mu}_j(t) &\convto_p \frac{ E(X_j \given O \geq t) P(O \geq t) }{ P(O \geq t) }\\
	&= E( X_j \given O \geq t)
\end{align}
as required. Similar arguments yield the results that $\hat{\sigma}_j^2(t) \convto_p Var(X_j \given O \geq t)$, $\hat \theta_j(t) \convto_p E(X_j \given O < t, \delta=1)$, and $\hat \psi^2_j(t) \convto_p Var(X_j \given O < t, \delta=1)$. Since $\{O < t, \delta = 1\} = \{ \min(T,C) < t, T \leq C \} = \{ T < t \}$, the desired results $\hat \theta_j(t) \convto_p \theta_j(t)$ and $\hat \psi^2_j(t) \convto_p \psi^2_j(t)$ are immediate. We therefore focus on $\hat \mu_j(t)$ and $\hat \sigma^2_j(t)$.

Having shown that $\hat{\mu}(t) \convto_p E(X \given O \geq t)$, we now show that $E(X \given O \geq t) = E(X \given T \geq t) \equiv \mu(t)$ under the above Conditions. Using $P(\cdot)$ to denote the density function, we have:

\begin{align*}
P(X_j \given O \geq t) &= P(X_j \given \min(T,C) \geq t)\\
	&= P(X_j \given T \geq t, C \geq t)\\
	&= \frac{P(T \geq t, C \geq t \given X_j) P(X_j)}{\int_X P(T \geq t, C \geq t \given X) P(X)}\\
	&= \frac{P(T \geq t \given X_j) P(C \geq t \given X_j) P(X_j)}{\int_{X} P(T \geq t \given X) P(C \geq t \given X) P(X) } \ \ \ \text{by Condition (1).}\\
	&= \frac{P(C \geq t) P(T \geq t \given X_j) P(X_j)}{P(C \geq t) \int_X P(T \geq t \given X) P(X) } \ \ \ \text{by Condition (2).}\\
	&= \frac{P(T \geq t \given X_j) P(X_j)}{P(T \geq t)}\\
	&= P(X_j \given T \geq t)
\end{align*}
Hence, we have the desired result on the expectation of $X_j$, yielding the consistency of $\hat \mu_j(t)$. The same argument gives the result on the expectation of $X_j^2$, yielding the consistency of $\hat \sigma^2_j(t)$.

\end{document}